# TOPOLOGICAL DESIGN OF AN ASYMMETRIC 3-TRANSLATIONAL PARALLEL MECHANISM WITH ZERO COUPLING DEGREE AND MOTION DECOUPLING

**Huiping Shen**
School of Mechanical Engineering, Changzhou University, Changzhou 213016, China
shp65@126.com

**Chengqi Wu**
School of Mechanical Engineering, Changzhou University, Changzhou 213016, China
753272704@qq.com

**Damien Chablat**
CNRS, Laboratoire des Sciences du Numérique de Nantes, UMR 6004
Nantes, France
Damien.Chablat@cnrs.fr

**Guang Lei Wu**
School of Mechanical Engineering, Dalian University of Science and Technology, Dalian 116031, China
gwu@dlut.edu.cn

**Ting-li Yang**
School of Mechanical Engineering, Changzhou University, Changzhou 213016, China
yangtl@126.com

## ABSTRACT

In this paper a new asymmetric 3-translational (3T) parallel manipulator, i.e., RPa(3R) 2R+RPa, with zero coupling degree and decoupled motion is firstly proposed according to topology design theory of parallel mechanism (PM) based on position and orientation characteristics (POC) equations. The main topological characteristics such as POC, degree of freedom and coupling degree are calculated. Then, the analytical formula for the direct and inverse kinematic are directly derived since coupling degree of the PM is zero. The study of singular configurations is simple because of the independence of the kinematic chains.

## INTRODUCTION

In many industrial production lines, process operations require pure translation movements only. Therefore, the 3-DOF translational parallel mechanism (TPM) has a significant potential value since it is a relatively simple structure and easily controlled [1].

Many scholars have being studied the TPM. For example: original design of 3-DOF TPM is the Delta Robot which was presented by Clavel [2]. The structure manipulators of TPM have been developed [3-5]. Tsai et al [6] presented the 3-DOF TPM, the moving actuators are prismatic joints and the sub-chain is 4R parallelogram mechanism (P is prismatic joint and R is revolute joint). The same architecture was optimized to have isotropic posture in the center of its workspace with 3 or 5 degrees of freedom in [7, 8]. Li et all developed a 3-UPU parallel mechanism (U is universal joint) [9] and analyzed the instantaneous kinematics performance of the TPM. In [10], the authors suggested a 3-RRC TPM and developed the forward and inverse solution equation (C is cylindrical joint). Considering the anisotropy of kinematics, Zhao et al [11] analyzed the dimension synthesis and kinematics of the 3-DOF TPM based on the Delta PM. Zeng et.al [12-14] introduced a 3-DOF TPM called as Tri-pyramid robot and presented a more detailed analytical approach for the Jacobi matrix. G. Bhutani et al [15] established a new design for the 3-UPU mechanism by taking into account mathematical models. Gao et al [16] developed a TPM with decoupled motion.

However, the most previous TPMs generally suffer from two major problems: i) the degree of coupling κ of the mechanism is greater than zero, which means its forward position solution is generally not analytical, and ii) the mechanism does not have input-output decoupling characteristics [17], which leads to the complexity of motion control and path planning.

According to topology design theory of PM based on position and orientation characteristics (POC) equations [18], a new TPM is proposed in this paper. The TPM is designed with simple structure and features zero coupling degree, which makes it possible to produce analytical models for the direct and inverse kinematic model. In addition, TPM's motion control and trajectory planning is easier thanks to its partially decoupled motion characteristics.

## STRUCTURAL DESIGN

The 3T parallel manipulator proposed in this paper is illustrated in Fig.1. The base platform 0 is connected to the moving platform 1 by two hybrid simple opened chain (HSOCs).

To illustrate the structure of the manipulator, the CAD design of the hybrid chain is shown in Fig. 1(A). Two HSOCs composed of series of links are shown in Fig. 1(B).

The first HSOC includes the branches I and II: The shorter link 3 of a parallelogram composed of four spherical pairs ( $S_a$ , $S_b$ , $S_c$ and $S_d$ ) is connected at point a, by actuated arm 2, to the base 0 by a revolute joint $R_{11}$. The extended part of the opposite link 3’ of the parallelogram is connected in parallel with a sub-chain composed of two links(6 and 5) and three parallel revolute joints (3R, i.e., $R_{23} \mathrm{P} R_{22} \mathrm{P} R_{21}$ ), which is denoted as $RP_a^{(4S)} 3R$. Further, two links (7 and 8) and two parallel revolute joints (2R, i.e., $R_{12} \| R_{13}$ ) are connected in series with the link 3’, which are connected to the moving platform 1. Because the two revolute joints(2R, i.e., $R_{11} \perp R_{21}$ ) of the base 0 are perpendicular to each other the HSOC can be recorded as $RP_a^{(4S)} 3R \perp 2\mathrm{R}$ , the end part of which is part of the moving platform 1 and produce three translations and one rotation.

The second HSOC includes the branch III: The parallelogram composed of four revolute joints ( $R_e$ , $R_f$ , $R_g$ and $R_h$ ) is connected in series with two parallel revolute joints (2R, i.e., $R_{32} \| R_{31}$ ), and is connected to the moving platform 1 by $R_{33} \left( R_{33} \| R_{32} \right)$ . The HSOC can be recorded as $RP_a^{(4R)}$ , the end part of which produce three translations and one rotation. It is obvious that the HSOC has the same branch as the typical Delta (but the Delta contains three such complex HSOCs).

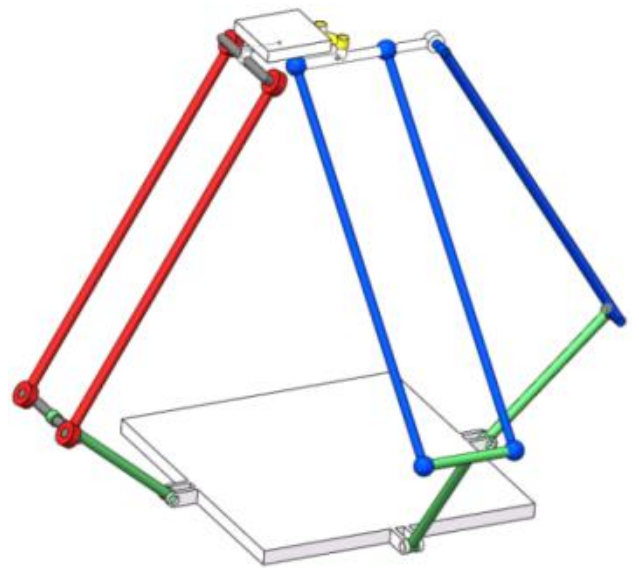

(A) 3D CAD DESIGN

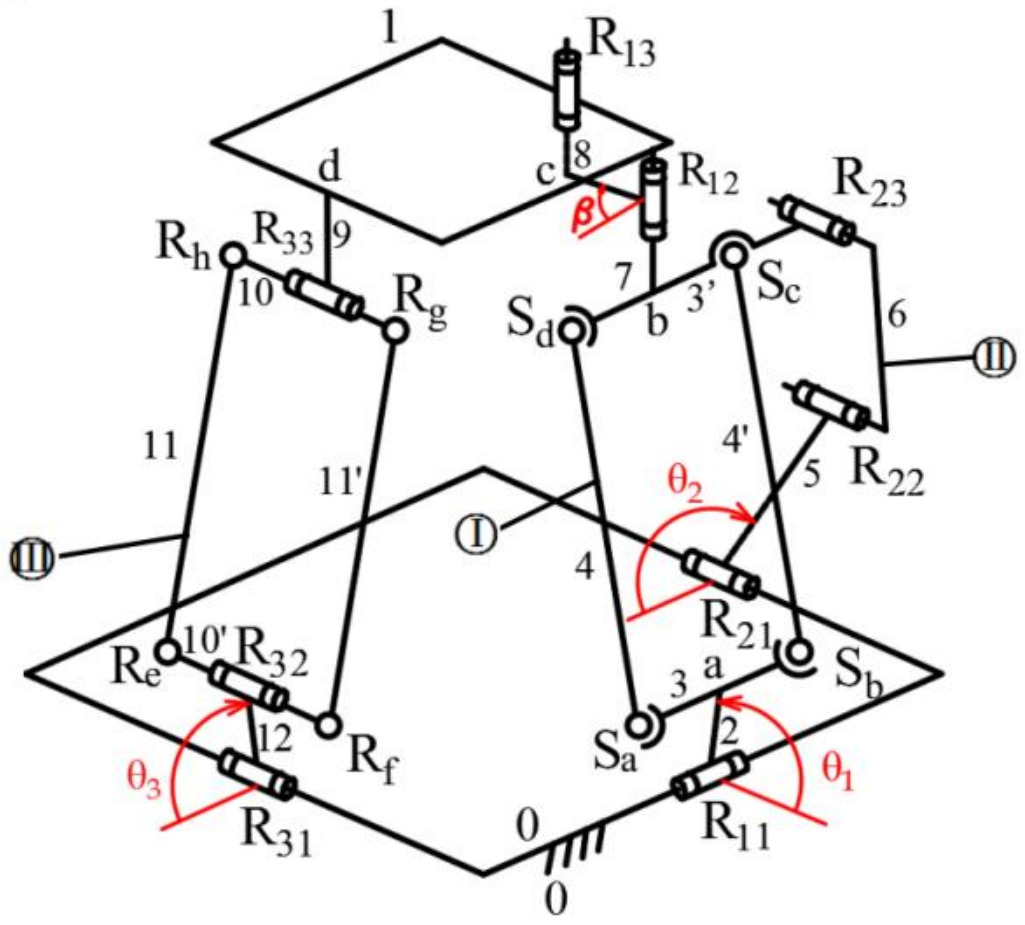

(B) KINEMATIC STRUCTURE

FIGURE 1. 3D CAD DESIGN AND KINEMATIC STRUCTURE OF 3T ROBOT

## TOPOLOGICAL ANALYSIS

### Analysis of the POC set

The POC set equations for serial and parallel mechanisms are expressed respectively as follows:

$$M_{bi} = \bigcup_{i=1}^{m} M_{Ji} \tag{1}$$

$$M_{Pa} = \bigcap_{i=1}^{n} M_{bi} \tag{2}$$

where

$M_{Ji}$ - POC set generated by the $i$-th joint;

$M_{bi}$ - POC set generated by the end link of $i$-th branched chain;
$M_{Pa}$ - POC set generated by the moving platform of PM.

The topological architecture of the two *HSOCs* of the mechanism can be denoted as:

① $HSOC_1\{R_{11}(-P^{4S}-P^{4S}-R^{4S}-R^{4S})-(R_{23}\parallel R_{22}\parallel R_{21})\perp(R_{12}\parallel R_{13})\}$
② $HSOC_2\{R_{31}(P^{(4R)})PR_{32}\,PR_{33}\}$.

The POC sets of the end of the two *HSOCs* are determined according to Eqs.(1) and (2) as follows:

$$M_{HSOC_1}=\left(\begin{bmatrix}t^1(\perp R_{11})\\ r^1(PR_{11})\end{bmatrix}\cup\begin{bmatrix}t^1(P\langle\rangle(abcd))\cup t^1(\perp(ab))\\ r^1(P(ab))\cup r^1(\perp(bd))\end{bmatrix}\right)$$
$$\cap\begin{bmatrix}t^2(\perp R_{23})\\ r^1(PR_{21})\end{bmatrix}\cup\begin{bmatrix}t^2(\perp R_{12})\\ r^1(PR_{12})\end{bmatrix}=\begin{bmatrix}t^3\\ r^1(PR_{12})\end{bmatrix}$$

$$M_{HSOC_2}=\begin{bmatrix}t^1(\perp R_{31})\\ r^1(PR_{31})\end{bmatrix}\cup\begin{bmatrix}t^1(\perp R_{32})\\ r^1(PR_{32})\end{bmatrix}\cup\begin{bmatrix}t^1(\perp R_{33})\\ r^1(PR_{33})\end{bmatrix}\cup\begin{bmatrix}t^1(PR_{32})\\ r^0\end{bmatrix}$$
$$=\begin{bmatrix}t^3\\ r^1(PR_{31})\end{bmatrix}$$

The POC set of the moving platform of this PM is determined from Eq.(2) by

$$M=M_{HS0C_1}\cap M_{HSOC_2}=\begin{bmatrix}t^3\\ r^0\end{bmatrix}$$

So, the moving platform 1 of the PM produces a pure translation motion.

**Determining the DOF**

The general and full-cycle DOF formula for PMs proposed in author's work[18] is given:

$$F=\sum_{i=1}^{m}f_i-\sum_{j=1}^{v}\xi_{Lj} \tag{3a}$$

$$\sum_{j=1}^{v}\xi_{Lj}=\dim\left\{\left(\bigcap_{i=1}^{j}M_{b_i}\right)\cup M_{b(j+1)}\right\} \tag{3b}$$

where, $F$ - DOF of PM. $f_i$- DOF of the $i^{th}$ joint. $m$ - number of all joints of the PM. $v$ - number of independent loops ($v=m-n+1$, $n$ - number of links). $\xi_{Lj}$ - number of independent equations of the $j^{th}$ loop. $\bigcap_{i=1}^{j}M_{b_i}$ - POC set generated by the sub-PM formed by the former $j$ branches; $M_{b(j+1)}$ - POC set generated by the end link of $j$+1sub-chains.

The mechanism can be broken down into two independent loops, and their constraint equations are calculated as follows:

① The first independent loop is consisted of branch I and II, the $HSOC_1$ is deduced as:

$$HSOC_1\{R_{11}(P^{4S}-P^{4S}-R^{4S}-R^{4S}-R_{23}\parallel R_{22}\parallel R_{21}\}$$

Thus, as obtained from Eq. (2), the POC set of the sub PM is

$$M_{Pa(1-2)}=M_I\cap M_{II}=\dim\left\{\begin{bmatrix}t^2\\ r^2\end{bmatrix}\cap\begin{bmatrix}t^2(\perp R_{21})\\ r^1(PR_{21})\end{bmatrix}\right\}=\begin{bmatrix}t^2\\ r^0\end{bmatrix} \tag{4}$$

In accordance with $Eq.$ (3), the independent constraint equation numbers $\xi_{L_1}$ and the DOF, respectively, can be obtained as follows:

$$\zeta_{L_1}=\dim\{M_I\cup M_{II}\}$$
$$=\dim\left\{\begin{bmatrix}t^2\\ r^2\end{bmatrix}\cup\begin{bmatrix}t^2(\perp R_{21})\\ r^1(PR_{21})\end{bmatrix}\right\}$$
$$=\dim\left\{\begin{bmatrix}t^3\\ r^3\end{bmatrix}\right\}=6$$

$$F=\sum_{i=1}^{m}f_i-\sum_{j=1}^{1}\zeta_{Lj}=8-6=2\,.$$

It can be seen that the output link 3' of the sub PM produces two dimensional translational motions in the *xoz* plane, and is only determined by the active joints $R_{11}$ and $R_{21}$. Therefore, the mechanism has partial motion decoupling.

② The above-mentioned sub PM and the branch chain III comprise the second loop:

$$HSOC_2\{R_{31}(P^{(4R)})\|R_{32}\|R_{33}\text{-}R_{13}\parallel R_{12}\}$$

In accordance with Eq. (3), the independent displacement equation numbers $\xi_{L_2}$ and the DOF, respectively, can be obtained as follows:

$$\xi_{L_2}=\dim\left\{\begin{bmatrix}t^2(\perp R_{12})\\ r^1(PR_{13})\end{bmatrix}\cup M_{III}\right\}=\dim\left\{\begin{bmatrix}t^2(\perp R_{12})\\ r^1(PR_{13})\end{bmatrix}\cup\begin{bmatrix}t^3\\ r^1(PR_{31})\end{bmatrix}\right\}$$
$$=\dim\left\{\begin{bmatrix}t^3\\ r^2(P\langle\rangle(R_{13},R_{31}))\end{bmatrix}\right\}=5$$
$$F=\sum_{i=1}^{m}f_i-\sum_{j=1}^{1}\zeta_{L_j}=(8+6)-(6+5)=3\,.$$

**Determining the coupling degree**

*Definition of coupling degree*

According to the composition principle of mechanism based on SOC units, any PM can be decomposed into a series of Assur kinematics chains (AKCs), and an AKC with $v$ independent loops can be broken down into $v$ single-open-chains (SOC). The constraint of the $j^{th}$ SOC is defined by

$$\Delta_j=\sum_{i=1}^{m_j} f_i - I_j - \zeta_{L_j} = \begin{cases} \Delta_j^- = -5,-4,-2,-1 \\ \Delta_j^0 = 0 \\ \Delta_j^+ = +1,+2,+3,... \end{cases} \quad (4)$$

where, $m_j$- number of joints contained in the $j^{th}$ $SOC_j$. $f_i$ - DOF of the $i^{th}$ joints; $I_J$ - number of actuated joints in the $j^{th}$ $SOC_j$; $\zeta_{L_j}$ - number of independent equations of the $j^{th}$ loop.

Then, the coupling degree of AKC is

$$\kappa = \frac{1}{2}\min\left\{\sum_{j=1}^{v}\left|\Delta_j\right|\right\} \quad (5)$$

The physical meaning of the coupling degree $\kappa$ can be explained as follows:

① The coupling degree $\kappa$ reflects the correlation and dependence between kinematic variables of each independent loop of the mechanism. It has been proved that the higher $\kappa$, the greater the complexity of the kinematic and dynamic problems of the mechanism will be.

② For $\kappa=0$, the motion of each loop can be obtained independently, and we can finally obtain the solution of the direct kinematic model; If $\kappa>0$, this means that the direct kinematic model must be solved using several constraint equations.

*Calculation of coupling degree*

The independent displacement equations $\xi_{Li}(i=1,2)$ of loop 1 and loop 2 have been calculated in the previous section, i.e., $\xi_{L_1}=6$, $\xi_{L_2}=5$, thus, the both coupling degrees are calculated by Eq. (4), respectively

$$\Delta_1=\sum f_i - I_j - \zeta_{L_j} = (5+3)-2-6=0$$

$$\Delta_2=\sum f_i - I_j - \zeta_{L_j} = 6-1-5=0$$

By Eq.(5), it can be seen that the mechanism contains two AKCs with zero coupling degree ($\kappa_1$=0, $\kappa_2$=0), hence, the closed-form forward position solution of the PM can be easily obtained by $AKC_1$ and $AKC_2$.

## POSITION ANALYSIS

### Establishment of the coordinate system and parameterization

To simplify reading, the mechanism shown in Fig. 1 is developed under a planar view in the Fig. 2. The base 0 and the moving platform 1 are square shaped, and the sizes of their sides are $2l_1$ and $2l_2$, respectively. The three revolute joints (3R, i.e., $R_{11}$, $R_{21}$ and $R_{31}$) of the base 0 are distributed at the midpoint of each edge.

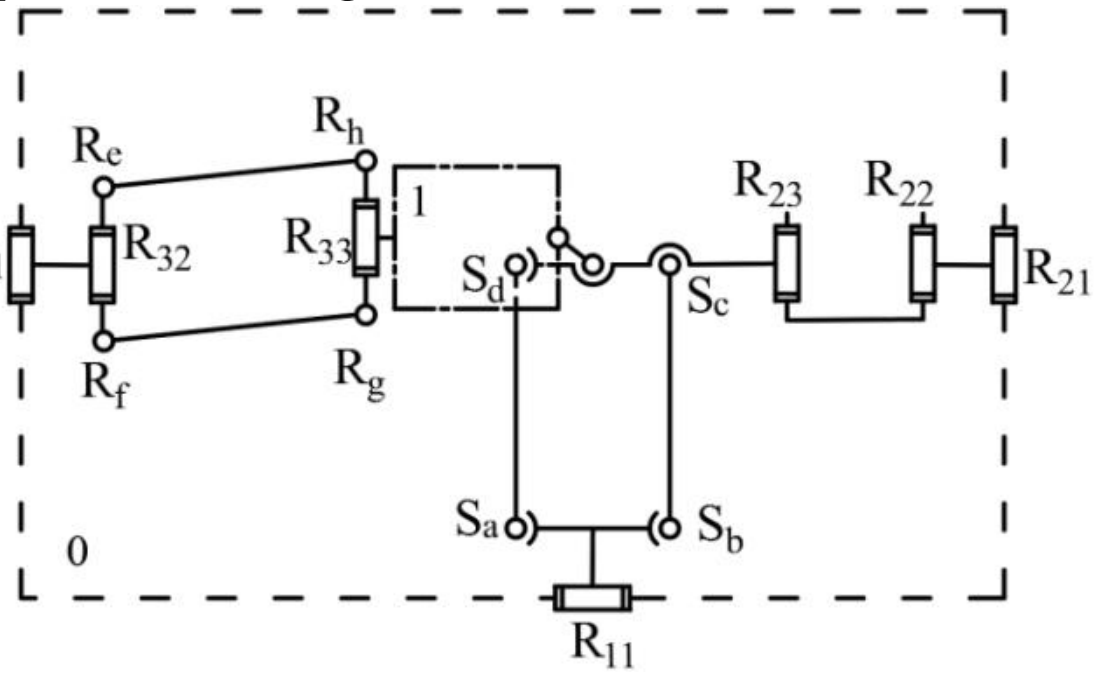

FIGURE 2. A PLANAR VIEW OF THE 3 ROBOTS

Without losing the generality, the frame coordinate system O-*xyz* is established on the base 0. The $x$-axis is parallel to the axis of $R_{11}$, and the $y$-axis is perpendicular to the axis of $R_{11}$.

On the moving platform 1, the moving coordinate system *uvw* is established at $P$. The $u$ axis and the $v$ axis are perpendicular and parallel to the axis of the $R_{33}$ respectively. Both $z$ and $w$ axis are determined by the right hand Cartesian coordinate rule, as shown in Fig. 3(A). For ease of comprehension, the 3T PM is redesigned in 3D view as shown in Fig. 3(B).

The length of each actuated arms 2, 5 and 12 is

$A_1B_1=A_2B_2=A_3B_3=l_3$ with $l_3\neq l_1$

The length of driven links 6 and the length of longer links of the both parallelograms 4 and 11 are

$B_1C_1=B_2C_2=B_3C_3=l_4$

The shorter links of the parallelogram are $2l_5$ and the point $B_i$ and $C_i$ ($i$=1, 2, 3) are the midpoint of the short edge. Thus, the parameters of the other links are respectively defined below.

$C_2D_2=l_5$, $C_1D_1=C_3D_3=E_1F_1=l_6$, $D_1E_1=l_7$.

The three input angles are defined as $\theta_1$, $\theta_2$ and $\theta_3$, as shown in Fig. 3(a). That is, the angle between the vectors $A_1B_1$ and the $y$-axis is $\theta_1$. The angles between the vectors $A_2B_2$, $A_3B_3$, and the $x$-axis are $\theta_2$ and $\theta_3$ respectively. The angle between the vectors $D_1E_1$ and the $x$-axis is β.

**Direct kinematics**

To solve the direct kinematic problem, i.e., to compute the position of the moving platform, we set the values of the actuated joints $\theta_1$, $\theta_2$ and $\theta_3$.

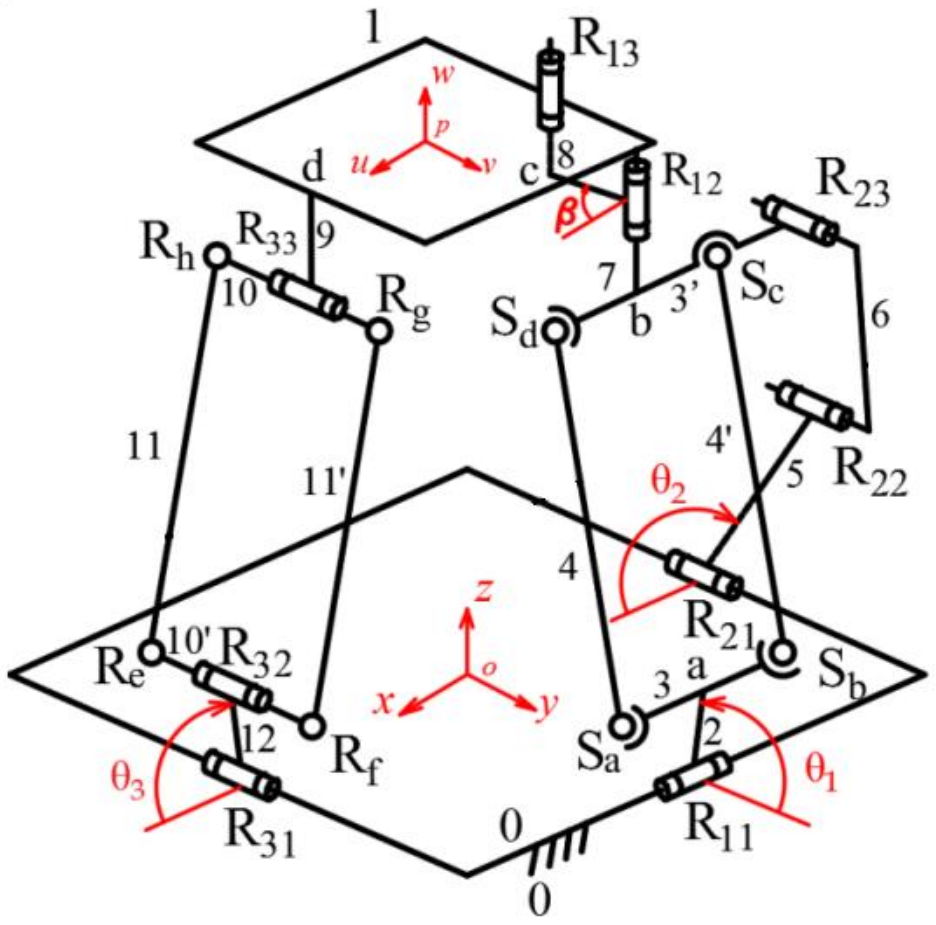

(A) KINEMATIC STRUCTURE

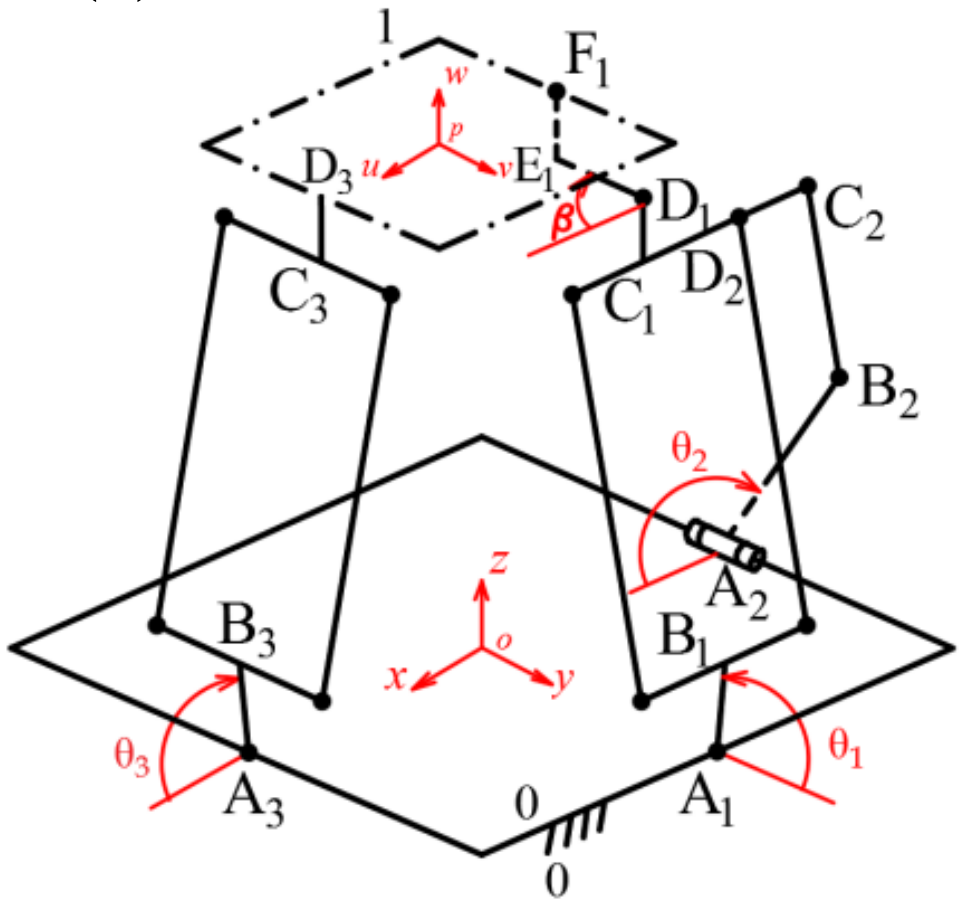

(B) KINEMATIC MODELING

FIGURE 3. PARAMETERIZATIONS OF THE 3T PM

*Direct kinematics of AKC$_1$*

The coordinates of $A_1$, $A_2$ and $A_3$ on the base platform 1 are $A_1=[0 \quad l_1 \quad 0]^T$, $A_2=[-l_1 \quad 0 \quad 0]^T$ and $A_3=[l_1 \quad 0 \quad 0]^T$.

The coordinates of each end-point of the three actuated arms 2, 5 and 12, i.e., $B_1$, $B_2$ and $B_3$ are easily calculated as

$$B_1=[0 \quad l_1+l_3\cos\theta_1 \quad l_3\sin\theta_1]^T,$$

$$B_2=[-l_1+l_3 cos\theta_2 \quad 0 \quad l_3\sin\theta_2]^T,$$

$$B_3=[l_1+l_3 cos\theta_3 \quad 0 \quad l_3\sin\theta_3]^T$$

As stated in Eq. (4), the output link 3' of the sub-PM can only produce two dimensional translational motions in the *xoz* plane, i.e. $y_{C_1}=y_{C_2}=0$. Due to the link length constraints defined by $B_1C_1=B_2C_2=l_4$, there are two constraint equations as below.

$$\begin{cases}\left(x_{C_1}-x_{B_1}\right)^2+\left(y_{C_1}-y_{B_1}\right)^2+\left(z_{C_1}-z_{B_1}\right)^2=l_4^2\\ \left(x_{C_2}-x_{B_2}\right)^2+\left(y_{C_2}-y_{B_2}\right)^2+\left(z_{C_2}-z_{B_2}\right)^2=l_4^2\end{cases} \tag{6}$$

Equation (6) leads to:

$$a_1x_{C_1}+b_1z_{C_1}=c_1 \tag{7}$$

Where

$a_1=2(x_{B_2}+2l_5)$, $b_1=2(z_{B_2}-z_{B_1})$,

$c_1=(x_{B_2}+2l_5)^2+z_{B_2}^2-y_{B_1}^2-z_{B_1}^2$

If $a_1$=0 and $b_1$=0, then $c_1=-y_{B_1}^2=0$. But $y_{B1}$ could not be zero. Hence, $a_1$ and $b_1$ are not zero at the same time, and we can have two cases as follows:

① $a_1=0$. In such a case, we have

$$\begin{cases}z_{C_1}=c_1/b_1\\ x_{C_1}=\pm\sqrt{l_4^2-(z_{B_1}-z_{C_1})^2-y_{B_1}^2}\end{cases} \tag{8a}$$

② $a_1\neq 0$. In such a case, we have

$$\begin{cases}z_{C_1}=\dfrac{e_1\pm\sqrt{e_1^2-4d_1f_1}}{2d_1}\\ z_{C_1}=\dfrac{c_1-b_1z_{C_1}}{a_1}\end{cases} \tag{8b}$$

where

$d_1=a_1^2+b_1^2$, $e_1=2(b_1c_1+z_{B_1}a_1^2)$, $f_1=a_1^2(y_{B_1}^2+z_{B_1}^2-l_4^2)+c_1^2$.

*Direct kinematics of AKC$_2$*

From the Fig. 3, the coordinates of $D_1$, $E_1$, $F_1$, $D_3$, and $C_3$ are defined as:

$$D_1=\left[x_{C_1} \quad 0 \quad z_{C_1}+l_6\right]^T,$$

$$E_1=\left[x_{D_1}+l_7cos\beta \quad l_7sin\beta \quad z_{D_1}\right]^T,$$

$$F_1=\left[x_{D_1}+l_7cos\beta \quad l_7sin\beta \quad z_{D_1}\right]^T,$$

$$D_3=\left[x_{D_1}+l_7cos\beta+2l_2 \quad l_7sin\beta \quad z_{D_1}\right]^T,$$

$$C_3=\left[x_{D_3} \quad y_{D_3} \quad y_{D_3}-l_6\right]^T.$$

Due to the link length constraints defined by $B_3C_3 = l_4$, the constraint equation can be deduced as below.

$$\left(x_{B_3} - x_{C_3}\right)^2 + \left(y_{B_3} - y_{C_3}\right)^2 + \left(z_{B_3} - z_{C_3}\right)^2 = l_4^2 \quad (9)$$

Equation (9) leads to

$$a_2 cos\beta = b_2 \quad (10)$$

With

$$a_2 = 2l_7\left(x_{D_1} - x_{B_3} + 2l_2\right),$$

$$b_2 = l_4^2 - l_7^2 - (x_{D_1} - x_{B_3} + 2l_2)^2 - (z_{D_1} - z_{B_3})^2$$

After $\beta$ is obtained by Eq. (10), the coordinate of $D_3$ and $F_1$ can be obtain. Thus, the coordinates of $P$ on the platform is

$$\begin{cases} x = \left(x_{D_3} + x_{F_1}\right)/2 \\ y = \left(y_{D_3} + y_{F_1}\right)/2 \\ z = \left(z_{D_3} + z_{F_1}\right)/2 \end{cases} \quad (11)$$

**Inverse kinematics**

To solve the inverse kinematics, we compute the values of $\theta_1$, $\theta_2$ and $\theta_3$ as a function of the coordinate $P$ of the moving platform. We have also to evaluate the value of β, which is a passive variable form the Eq. (10).

*Solution for* $\beta$

For a given position of the moving platform, there are two positions for points $C_1$ and $C_2$ with

$$\beta = \arcsin\left(\frac{y_{D_3}}{l_7}\right) \quad \text{or} \quad \beta = \pi - \arcsin\left(\frac{y_{D_3}}{l_7}\right)$$

For each position of $C_1$ and $C_2$, we can calculate $\theta_1$ and $\theta_2$.

*Solution for* $\theta_1$ *and* $\theta_2$*:*

From Fig. 3, the coordinates of $F_1$, $E_1$ and $D_1$ are

$$F_1 = [x - l_2, y, z]^T, \quad E_1 = [x - l_2, y, z - l_6]^T, \quad D_1 = \left[x_{D_1}, 0, z - l_6\right]^T.$$

Due to the link length constraints defined by $D_1E_1 = l_7$, the coordinates of $D_1$, $C_1$ and $C_2$ can be deduced as below:

$$D_1 = \left[x - l_2 + \sqrt{l_7^2 - y^2} \quad 0 \quad z - l_6\right]^T$$

$$C_1 = \left[x_{D_1} \quad 0 \quad z_{D_1} - l_6\right]^T, \quad C_2 = \left[x_{D_1} - 2l_5 \quad 0 \quad z_{D_1} - l_6\right]^T.$$

Then, due to the link length constraints defined by $B_1C_1 = B_2C_2 = l_4$, i.e.

$$\begin{cases} \left(x_{C_1} - x_{B_1}\right)^2 + \left(y_{C_1} - y_{B_1}\right)^2 + \left(z_{C_1} - z_{B_1}\right)^2 = l_4^2 \\ \left(x_{C_2} - x_{B_2}\right)^2 + \left(y_{C_2} - y_{B_2}\right)^2 + \left(z_{C_2} - z_{B_2}\right)^2 = l_4^2 \end{cases} \quad (12)$$

Then, Eq. (12) leads to:

$$\theta_i = 2\arctan\left((2l_3z_i \pm \sqrt{4z_i^2l_3^2 - g_ih_i})/g_i\right), \text{ for } i=1, 2 \quad (13)$$

where

$z_1 = z_{C_1}$, $z_2 = z_{C_2}$, $h_1 = (l_1 - l_3)^2 - l_4^2 + x_{C_1}^2 + z_{C_1}^2$,

$g_1 = (l_1 + l_3)^2 - l_4^2 + x_{C_1}^2 + z_{C_1}^2$,

$h_2 = (l_1 - l_3)^2 - l_4^2 + x_{C_1}^2 + z_{C_1}^2 + 2x_{C_2}(l_1 - l_3)$,

$g_2 = (l_1 + l_3)^2 - l_4^2 + x_{C_1}^2 + z_{C_1}^2 + 2x_{C_2}(l_1 + l_3)$.

*Solution for* $\theta_3$

Similarly, the coordinates of $D_3$ and $C_3$ can be easily obtained

$$D_3 = [x + l_2, y, z]^T, \quad C_3 = [x + l_2, y, z - l_6]^T.$$

Due to the link length constraints defined by $B_3C_3 = l_4$, the constraint equation is established as below.

$$\left(x_{C_3} - x_{B_3}\right)^2 + \left(y_{C_3} - y_{B_3}\right)^2 + \left(z_{C_3} - z_{B_3}\right)^2 = l_4^2 \quad (14)$$

From Eq. (14), we can evaluate $\theta_3$ as following

$$\theta_3 = 2\arctan\left((2l_3z_3 \pm \sqrt{4z_3^2l_3^2 - g_3h_3})/g_3\right) \quad (15)$$

Where

$z_3 = z_{C_3}$, $g_3 = (l_1 - l_3)^2 - l_4^2 + y_{C_3}^2 + z_{C_3}^2$,

$h_3 = (l_1 + l_3)^2 - l_4^2 + y_{C_3}^2 + z_{C_3}^2$

We can conclude that this robot has 16 solutions to the inverse kinematic model, twice as many as the Delta robot. This is due to the mobile platform which is made with two parts. For the same position of $P$, there are two values for $\beta$ and for each $\beta$ value, there are two values for $\theta_1$ and $\theta_2$. However, there are only two $\theta_3$ values in total. The number of solutions in the inverse kinematic model is the product of 4 by 4 by 2, i.e. 16 solutions.

## SINGULARITY ANALYSIS

The singularity analysis of parallel robots has been well-documented in the literature. We can find the parallel and serial Jacobian matrix, named A and B respectively [19, 20], by differentiating the constraint equations with respect to time.

Then we obtain the parallel and serial singularities by studying det(A) and det(B) , respectively [21].

For the mechanism studied, the singularities are similar to those of a Delta robot. A new singular configuration exists because the mobile platform is made in two parts. When β is equal to $\pi/2$ or $3\pi/2$, the robot admit a new singular configuration because the determinant of the parallel Jacobian matrix A vanished. This defines in the workspace, a vertical plane passing through the origin and parallel to the plane (*xz*).

## CASE STUDY

To illustrate this study, we refer to the dimension parameters of the ABB robot 14R, i.e, $l_1=300$, $l_2=70$, $l_3=350$, $l_4=800$, $l_5=100$, $l_6=10$ and $l_7=50$. Its parallel and serial singularity can be computed with the Siropa library [22]. The determinant of matrix A can be written

$$\sin(\beta)\left(35\cos(\theta3)-x_{C_1}+16\right)x_{C_1}\left(35\sin(\theta2)-z_{C_1}\right)=0 \qquad (16)$$

and that of matrix B as

$$\left(-2100\,S_{\theta_1}-70\,z_{C_1}\,C_{\theta_1}\right)\left(1400\,S_{\theta_2}-70\,z_{C_1}\,C_{\theta_2}\right)\left(350\,C_\beta S_{\theta_3}+70\,x_{C_1}\,S_{\theta_3}-1120 S_{\theta_3}-70\,z_{C_1}\,C_{\theta_3}\right)C_\beta=0 \qquad (17)$$

By using Groebner base elimination methods, these surfaces can be plotted in the Cartesian space. Figure 4 depicts the singular configurations in the Cartesian space and Fig. 5 the serial singularities with a different color for each term of the Eqs (15) and (16).

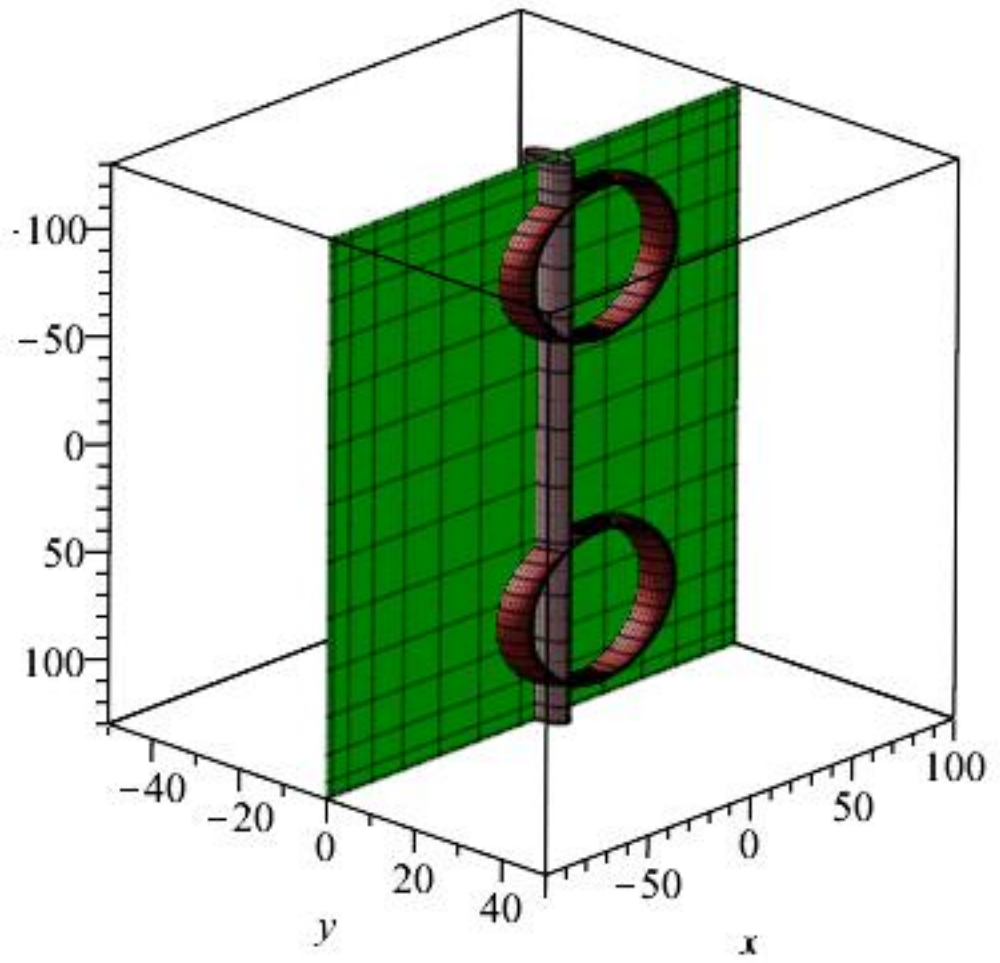

FIGURE 4. PARALLEL SINGULARITIESOF THE ROBOT

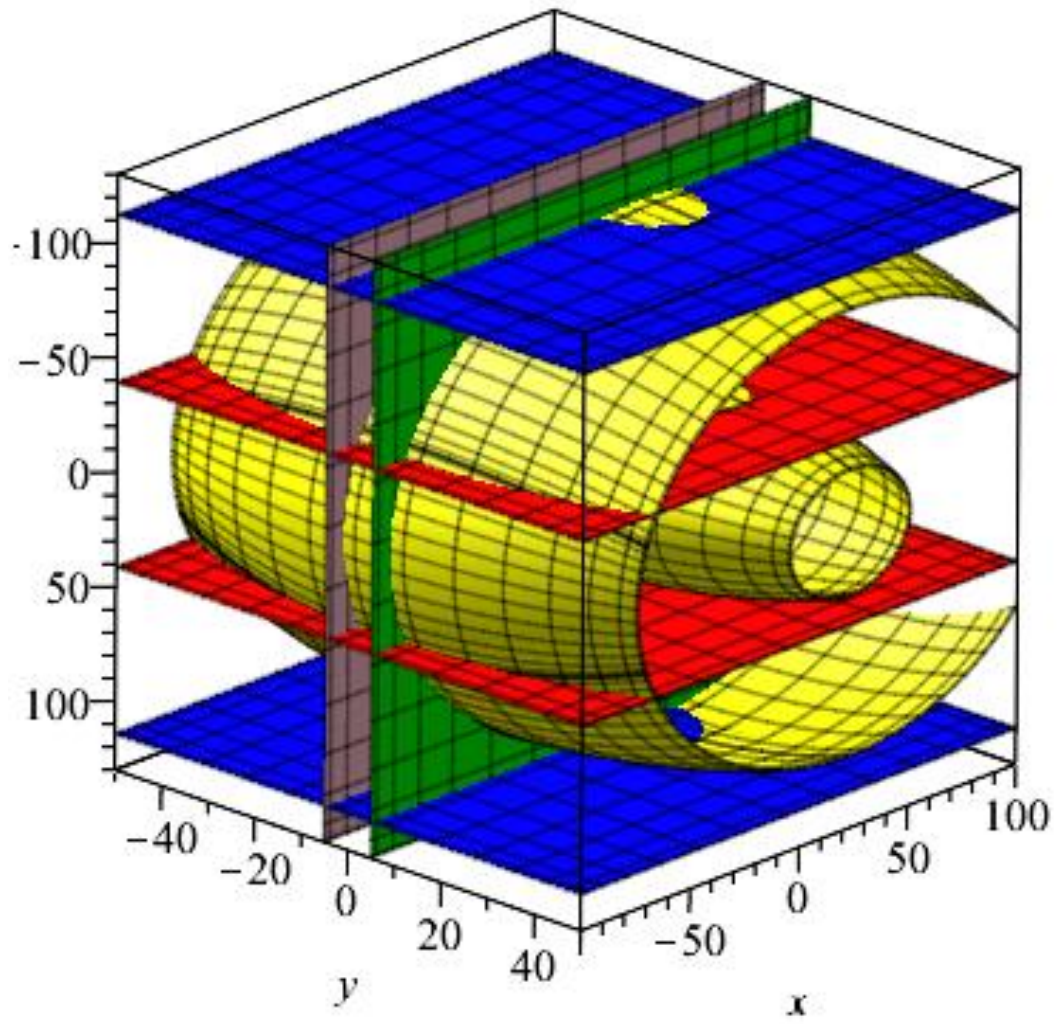

FIGURE 5. SERIAL SINGULARITIESOF THE ROBOT

## CONCLUSIONS

In this article, a new asymmetrical new parallel mechanism of pure parallel translation with a degree of zero coupling and complete motion decoupling has been proposed. The inverse and direct kinematic models are obtained. The mechanism is simple in structure and easy to manufacture, which can be used in transportation, positioning and other operations of the manufacturing industry. The study of singular configurations is simple because of the independence of the kinematic chains.

## ACKNOWLEDGMENTS:

This research is sponsored by the NSFC (Grant No. 51475050 and No. 51375062) and Jiangsu Key Development Project (No.BE2015043).

## REFERENCES

[1] Liu X., Wang J., Kinematic and workspace analysis of a new type of spatial 3-DOF parallel mechanism. Journal of Mechanical Engineering, 2001, 37(10):36-39.

[2] Clavel R., A Fast Robot with Parallel Geometry // Proc. Int. Symposium on Industrial Robots. 1988:91-100.

[3] Stock M, Miller K, Optimal Kinematic Design of Spatial Parallel Manipulators: Application to Linear Delta Robot. Journal of Mechanical Design, 2003, 125(2):292-301.

[4] Bouri M, Clavel R., The Linear Delta: Developments and Applications // Robotics. VDE, 2010:1-8.

[5] Kelaiaia R, Company O, Zaatri A., Multiobjective optimization of a linear Delta parallel robot. Mechanism & Machine Theory, 2012, 50(2):159–178.

[6] Tsai L. W, Walsh G C., Stamper R E., Kinematics of a Novel Three DOF Translational Platform. 1996, 4:3446-3451.

[7] Chablat D. and Wenger P., Architecture Optimization of a 3-DOF Parallel Mechanism for Machining Applications, the Orthoglide, IEEE Transactions On Robotics and Automation, 19(3):403-410, June, 2003.

[8] Caro S., Chablat D., Lemoine P., Wenger P., Kinematic Analysis and Trajectory Planning of the Orthoglide 5-Axis, Proceedings of the ASME 2015, International Design Engineering Technical Conferences & Computers and Information in Engineering Conference, Aug 2015, 2015.

[9] Li S, Huang Z, Zuo R., Kinematics of a Special 3-DOF 3-UPU Parallel Manipulator // ASME 2002 International Design Engineering Technical Conferences and Computers and Information in Engineering Conference. 2002:1035-1040.

[10] Zhao T., Huang Z., Kinematics analysis of a three dimensional mobile parallel platform mechanism. China Mechanical Engineering, 2001, 12(6):612-616.

[11] Zhao Y., Dimensional synthesis of a three translational degrees of freedom parallel robot while considering kinematic anisotropic property. Robotics and Computer-Integrated Manufacturing, 2013, 29(1):169-179.

[12] Zeng Q, Ehmann K F, Cao J., Tri-pyramid Robot: Design and kinematic analysis of a 3-DOF translational parallel manipulator. Pergamon Press, Inc. 2014.

[13] Qiang Z., Ehmann K F., Jian C., Tri-pyramid Robot: stiffness modeling of a 3-DOF translational parallel manipulator. Robotica, 2016, 34(2):383-402.

[14] Lee S, Zeng Q, Ehmann K F., Error modeling for sensitivity analysis and calibration of the tri-pyramid parallel robot. International Journal of Advanced Manufacturing Technology, 2017(5):1-14.

[15] Bhutani G, Dwarakanath T A., Novel design solution to high precision 3 axes translational parallel mechanism. Mechanism & Machine Theory, 2014, 75(5):118-130.

[16] Li W., Gao F., Zhang J., A three-DOF translational manipulator with decoupled geometry. Cambridge University Press, 2005.

[17] Shen H., Xiong K., Kinematic decoupling design method and application of parallel mechanism .Trans. of The Chinese Society of Agricultural Machinery, 2016, 47(6):348-356.

[18] Yang T., Liu A., Shen H., et.al, Topology design of robot mechanism. Springer, 2018.

[19] Gosselin C., and Angeles J., Singularity analysis of closed-loop kinematic chains, in IEEE Transactions on Robotics and Automation, 6(3), pp. 281–290, June 1990.

[20] Wenger Ph., Chablat D., Definition sets for the direct kinematics of parallel manipulators, *8th International Conference in Advanced Robotics*, pp. 859-864, 1997.

[21] Chablat D., Wenger Ph., Working modes and aspects in fully-parallel manipulator, *Proceedings of IEEE International Conference on Robotics and Automation*, pp. 1964–1969, May 1998.

[22] Siropa, Algebraic and robotic functions, http://siropa.gforge.inria.fr/doc/files/siropa-mpl.html, 2018.